\documentclass[10pt,twocolumn,letterpaper]{article}

\usepackage[pagenumbers]{cvpr} 
\usepackage{graphicx}
\usepackage{amsmath}
\usepackage{amssymb}
\usepackage{booktabs}
\usepackage{amsmath}
\usepackage{makecell}
\usepackage{bm}
\usepackage{algorithm}
\usepackage{algpseudocode}
\usepackage{multirow}

\usepackage[pagebackref,breaklinks,colorlinks]{hyperref}

\usepackage[capitalize]{cleveref}
\crefname{section}{Sec.}{Secs.}
\Crefname{section}{Section}{Sections}
\Crefname{table}{Table}{Tables}
\crefname{table}{Tab.}{Tabs.}

\begin{document}

\title{Fully Self-Supervised Learning for Semantic Segmentation}

\author{Yuan Wang\\
Tsinghua University\\
ShenZhen, China\\
{\tt\small wangyuan19@mails.tsinghua.edu.cn}
\and
Wei Zhuo*\\
Tencent\\
ShenZhen, China\\
{\tt\small weizhuo@tencent.com}
\and
Yucong Li\\
Tencent\\
ShenZhen, China\\
{\tt\small 47523255@qq.com}
\and
Zhi Wang\\
Tsinghua University\\
ShenZhen, China\\
{\tt\small  wangzhi@sz.tsinghua.edu.cn}
\and
Qi Ju \\
Tencent\\
ShenZhen, China\\
{\tt\small damonju@tencent.com}
\and
Wenwu Zhu*\\
Tsinghua University\\
Beijing, China\\
{\tt\small  wwzhu@tsinghua.edu.cn}
}

\maketitle

\begin{abstract}

In this work, we present a fully self-supervised framework for semantic segmentation($FS^4$). A fully bootstrapped strategy for semantic segmentation, which saves efforts for the huge amount of annotation, is crucial for building customized models from end-to-end for open-world domains. This application is eagerly needed in realistic scenarios. Even though recent self-supervised semantic segmentation methods have gained great progress, these works however heavily depend on the fully-supervised pretrained model and make it impossible a fully self-supervised pipeline. To solve this problem, we proposed a bootstrapped training scheme for semantic segmentation, which fully leveraged the global semantic knowledge for self-supervision with our proposed PGG strategy and CAE module. In particular, we perform pixel clustering and assignments for segmentation supervision. Preventing it from clustering a mess, we proposed 1) a \textbf{pyramid-global-guided (PGG)} training strategy to supervise the learning with pyramid image/patch-level pseudo labels, which are generated by grouping the unsupervised features. The stable global and pyramid semantic pseudo labels can prevent the segmentation from learning too many clutter regions or degrading to one background region; 2) in addition, we proposed \textbf{context-aware embedding (CAE)} module to generate global feature embedding in view of its neighbors close both in space and appearance in a non-trivial way. We evaluate our method on the large-scale COCO-Stuff dataset and achieved \textbf{7.19 mIoU} improvements on both things and stuff objects.

\end{abstract}
\renewcommand{\thefootnote}{*}
\footnotetext{Corresponding authors. }
\renewcommand{\thefootnote}{}
\footnotetext{This work was done while Yuan Wang was a Research Intern in Tencent.}


\section{Introduction}
Semantic segmentation is a task that gives each pixel in one image a class label. Fully supervised segmentation has gained great success due to deep learning on massive annotations. Pixel-level annotation is however extremely expensive. It is merely impossible to annotate pixels for the increasing open-world applications. This inspires us to design a fully unsupervised semantic segmentation scheme that can automatically recognize pixels belonging to different classes without human annotations. 

\begin{figure}
  \centering
  \includegraphics[width=\linewidth]{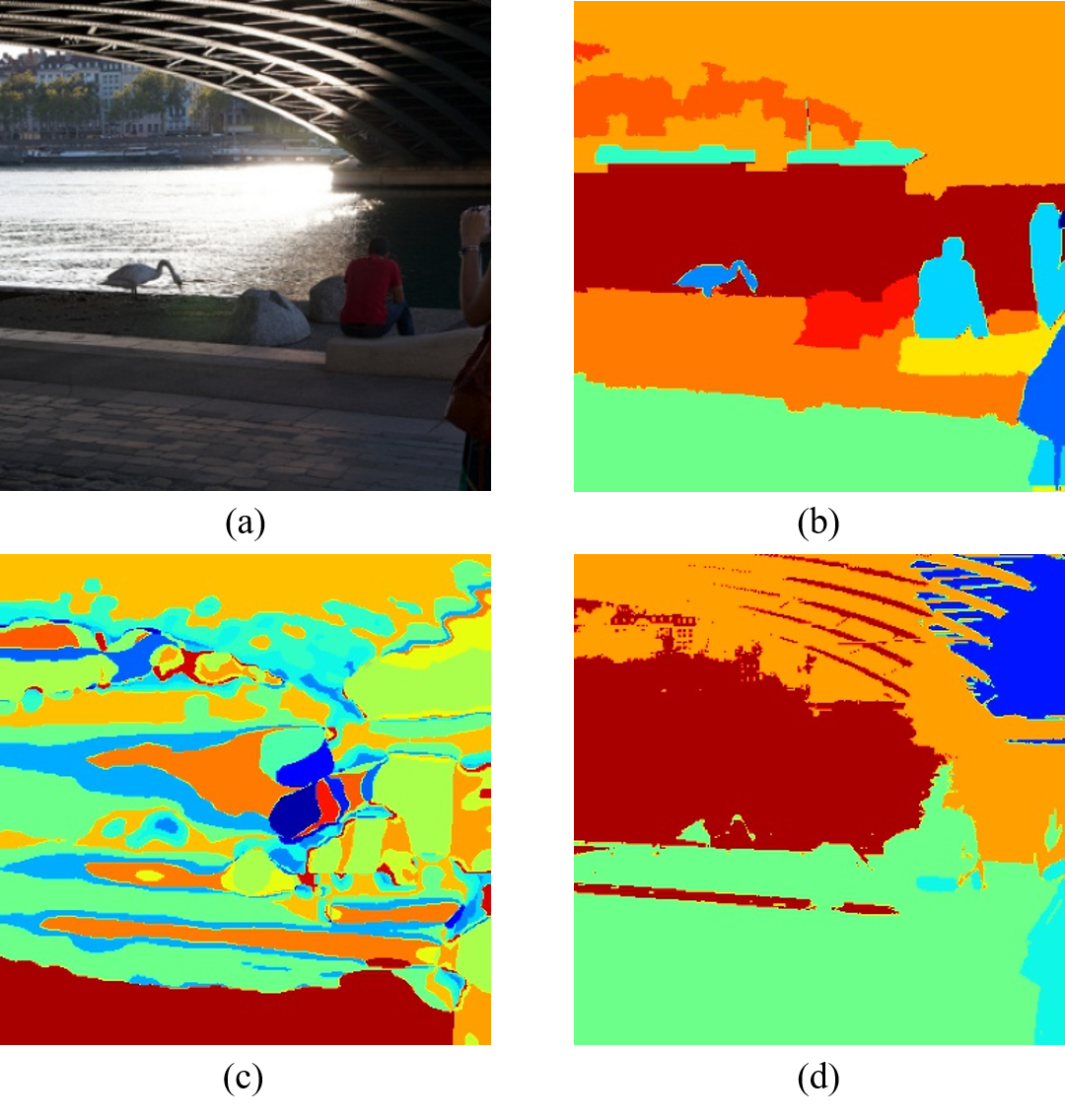}
  \caption{Example of segmentation results, where (a) is the original image, (b) is the ground truth, (c) is the result of PiCIE\cite{picie} and (d) is our result. We use the same color to visualize one semantic class in all above sub-images. Our method performs much better than the PiCIE conterpart, where we predict the correct semantic on river part while the PiCIE predicts a clutter of mess regions there.}
  \label{fig:introfig}
\end{figure}

\begin{figure*}
  \centering
  \includegraphics[width=\linewidth]{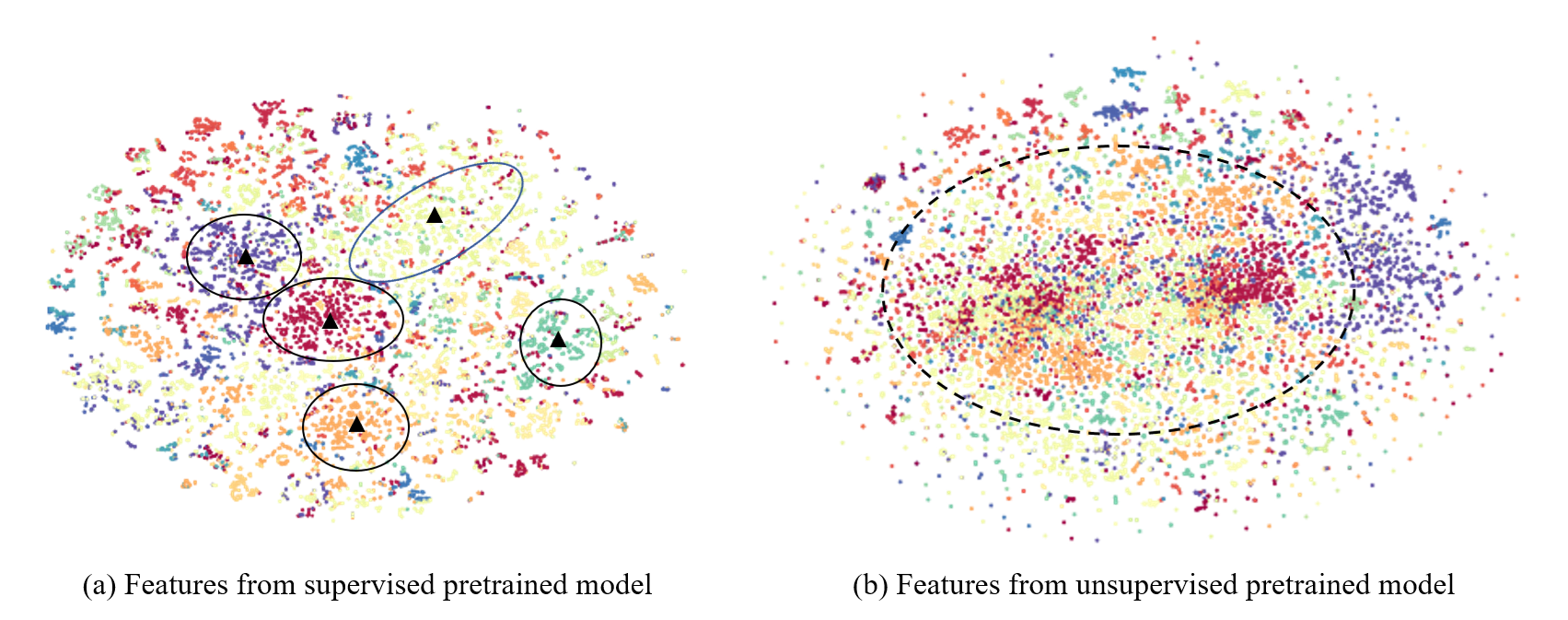}
  \caption{Pixel feature distribution on the supervised model\cite{imagenet} and unsupervised model\cite{swav} visualized using t-SNE. Here we grabbed the officially released pretrained models and extracted the features of their last layers on COCO-stuff dataset individually. We then resized them to size 80x80 using bilinear operation for pixel-wise grouping. We can see that the supervised pretrained model, i.e. ImageNet model, has better clusters on pixels. Since the whole learning is a bootstrapped process~\cite{iic, picie}, a good clustering at the initial state is crucial for globally guiding the subsequent pixel grouping. Better view in color.}
  \label{fig:intro_tsne}
\end{figure*}

The target of self-supervised semantic segmentation is to bootstrapped discover the categories of pixels. It requires the model can automatically discover existing semantic-meaningful categories in the dataset and group pixels to the corresponding categories.
Just recently, the pioneer works\cite{iic, picie} gave their attempts on this task by clustering on local elements, i.e., patches or pixels. These methods have achieved promising results based on ImageNet pretrained model, which is learned on 1k-class image annotations.  For short, we name the supervised ImageNet pretrained model as \textit{ImageNet model} afterwards. Altering the ImageNet model from an unsupervised pretrained model, which can be acquired by any instance-level self-supervised (ISS) methods such as \cite{swav,mocov2}, the performance drops to nearly half. The reason for this fact is analyzed in Figure~\ref{fig:intro_tsne}. The fully self-supervised capacity of segmentation, however, is important, since it is crucial for end-to-end customized model building for intensive open-world scenarios. In this work, we target at the \textit{fully} unsupervised semantic segmentation. To this end, we propose a novel method that fully makes use of the global knowledge and raises the performance twice the mIoU, and matches the results based on a well-supervised pretrained model.

To build a fully supervised pipeline, a straightforward way is to utilized an instance-level unsupervised method\cite{mocov2, swav} to provide an initial model and then perform local(pixel-level) bootstrapped clustering and learning on it, such as \cite{iic, picie}. Grouping based on pixel/local-patches features itself is however unstable. Due to lack of global image-level guidance, its segmentation is easy to have too many small regions or too large background/stuff regions that cross boundaries. Starting from an unsupervised pretrained model makes the issue more serious, as shown in Fig~\ref{fig:introfig}.

This could be because the local features of the supervised pretrained model are well-tuned for the object labels and its strong objective embedding in local features can work as global guidance during dataset clustering, while the local features from unsupervised methods are not well-tuned, as shown in Figure.~\ref{fig:intro_tsne}. Inspired by the above assumption, we propose a novel fully self-supervised segmentation method which fully leverages the global semantic knowledge with our pyramid global guidance(PGG) strategy for global supervision and our context-aware embedding(CAE) for global feature embedding.

In our work, we build a novel framework that performs two-level clustering, i.e. image-level and pixel-level, in two stages separately. The image-level clustering leverages the existent success of ISS methods and assigns image-level pseudo labels to pyramid views of images. The pseudo labels contain high-level semantics which is consistent in the dataset. The pyramid image pseudo labels keep unchanged during subsequent training on pixel-level clustering.  Here we need to mention that even though instance self-supervised methods, such as \cite{swav, mocov2, BYOL, simclrv2, simsiam, iic, deepcluster} lack capacity for direct dense recognition, such as pixel-wise segmentation, their global features is semantic meaningful, and they can work as the global signal for guidance.  We generate the labels on pyramids instead of one label for the whole image to provide finer supervision on images with multiple instances.

For the pixel-clustering stage, we follow \cite{picie} to bootstrap discover semantic clusters on pixel features and segment them. Improved on \cite{picie}, we 1) used the obtained pyramid image pseudo labels to supervise image labeling through a CAM module\cite{cam} on pixel labeling during the whole learning. In this way, our training intrinsically forces the pixel clustering to pay more attention to high-level semantics that defines the image class. This PGG strategy effectively bridges the gap raised by the suboptimal pretrained model; 2) In addition, we proposed a CAE module to enable the pixel clustering to be aware of its spatial neighbors in the image. We found that two close parts which are apparently the same semantic class depending on color can be segmented apart in pixel clustering of \cite{picie}, such as the river part in Figure~\ref{fig:introfig}. Inspired by this observation, we attempt to use raw image features for segmentation, such as the color and position cues used in classical segmentation, such as graph cut\cite{graphcut}, etc. The raw features, however, has been discovered to easily lead to collapsing solution in previous works. In our work, we found a non-trivial way to utilize both semantic features and raw features to improve the segmentation but avoid learning collapsing.  

Here we keep the above two-level clusters disjoint to avoid global instance features being infected by unstable local ones. 

In summary, our contributions are listed as follows, 
\begin{itemize}
    \item{We designed a novel and effective pipeline for the pioneer task of fully self-supervised semantic segmentation, which has great practical value for the widely open-world recognition scenarios.} 
    \item{We found a effective way using pyramid image-level pseudo-labels on the $FS^4$ task. The module we proposed is robust to process general images with multiple instances. } 
    \item{We proposed a novel context-aware embedding module that improve the features by both semantic features and raw image features, and found a non-trivial way to avoid collapse results while using raw features.}

\end{itemize}
Here note that our method does not require any kinds of labels. After training, the pixel cluster centers are used to segment the images in the validation dataset by assigning each pixel a label according to the distances between the pixel features and cluster centers. We show that in COCO-stuff \cite{stuff} dataset, our method can outperform the previous methods nearly \textbf{twice} on mIoU when both things and stuff parts are counted in.

\section{Related work}
\paragraph{Self-supervised learning.} self-supervised learning, or unsupervised learning in another name, develops very quickly recently. Most of these works focus on learning representations on an image level. Among these works, \cite{mocov2,simclrv2, simclrv2, simsiam, BYOL} learn the representations by forcing the models to learn instances from different images uniquely, and they have no cluster concepts. Some other works, such as \cite{swav,deepcluster}, try to introduce class semantics by clustering procedure. In this way, they naturally assign a pseudo label for each image. Note here, to avoid trivial solution, \cite{deepcluster} conducts offline clustering, and \cite{swav} uses the Sinkhorn-Knopp algorithm to prevent all images grouped to one large cluster. All the above methods can be categorized into instance-level self-supervised learning, which focus on distinguishing images. To this end, only the most discriminative parts in each image may be focused on, and many stuff information is ignored. This is contradicting with the targets of dense prediction, such as object detection and segmentation, which care about the boundaries of both stuff(background structures) and things(objects).  As a consequence, the instance self-supervised models can hardly segment an image directly. To make the unsupervised model adapted to dense downstream target, \cite{pixel-pro, contrast-conv, dense-contrast} designs contrast training strategy with dense correspondence. These methods, however, still cannot realize automatically recognize pixels.

\paragraph{Self-supervised semantic segmentation. } Very recently the self-supervised semantic segmentation has gained increasing attention\cite{iic,picie}. These methods aim to recognize classes of dense pixels. Clustering is a straightforward way to discover semantic classes and perform recognition on pixels or patches. IIC\cite{iic} an invariant information clustering method via maximizing the mutual information between encoded image pairs. A later work \cite{picie} conducts alternative offline pixel-wise clustering and online training, where the training is led by the invariance and equivariance objective on the assigned clusters. It also introduces latent contrast learning when the pixel is trained to be assigned to the cluster center/prototype which is assigned offline ahead.  Reference \cite{picie} however only depends on local features, and it performs much worse when the training starts from a sub-optimal point. Our method attempts to improve it via introducing image-level pseudo labels, which is related to weakly supervised learning.

\paragraph{Weakly supervised segmentation.} 
In a weakly supervised task, it utilizes weak labels of images to facilitate training supervision. Generally, these labels can be bounding boxes, image-level labels, etc. For the target of segmentation, class activation maps (CAM)\cite{cam}, which is the response map generated from a pretrained image classifier, are widely used as pseudo labels. Later works, such as \cite{seam,PMM}, are proposed to refine the CAM via leveraging the consistency among segments on different geometric or photometric transformations of the same image. They proved that when learning the refined pseudo labels we can get better segmentation. The existent weak-supervised segmentation methods depend on an image-level classifier, whose classes are usually defined by the foreground objects. In our case, the semantics of the image pseudo labels are mixed, which is bootstrapped popped up depending on the objects, background contents,   scene types and etc, through the learning and discovering.

\section{Baseline Method}
Before we step on our method, we first introduce a baseline method\cite{picie} that our method builds on. This method builds the unsupervised semantic segmentation learning by clustering pixel features based on their invariance and equivariance regulation. 
We will format the training procedure mathematically in the following part.

In the baseline work, it builds a siamese network with two branches.
For image $x$, on each branch, it applies random photometric transformations, such as Gaussian blur and color distortion, on the image independently and obtains an image view. On one branch, it first applies geometric transform such as cropping and flipping on its image view and then fed it to the convolutional neural network (CNN) model to get the feature set $F^{(1)}$. For the other branch, it first feeds the image view to CNN model and then performs random geometric transform to get the final feature set
$F^{(2)}$. 

We then apply K-means clustering method to the feature maps from the two branches seperately. On the two branches, it maintains two sets of clusters individually, that is, two sets of labels $Y^{(1)} = \{y_1^{(1)}, y_2^{(1)}, ...,y_N^{(1)}\}, Y^{(2)} = \{y_1^{(2)}, y_2^{(2)}, ..., y_N^{(2)}\}$ for each pixels and two sets of cluster centers $\bm{\mu}^{(1)} = \{\mu^{(1)}_1,\mu^{(1)}_2,...,\mu^{(1)}_K\}, \bm{\mu}^{(2)}= \{\mu^{(2)}_1,\mu^{(2)}_2,...,\mu^{(2)}_K\}$. According to the PiCIE assumption that \textit{pixel labels should be invariant to its color transformation and equivariance to its geometric transformation},  the cluster labels from two branches should be equivariance. The loss is formulated as following:
\begin{equation}
    L_{p} = L_{\text{within}} + L_{\text{cross}}
\label{Lp-loss}
\end{equation}
\begin{equation}
\begin{aligned}
    L_{\text{within}} = \frac{1}{N} &\sum_i \left ( L_{\text{clust}} (F^{(1)}_i,y_i^{(1)}, \bm{\mu}^{(1)}) \right.\\
    &\left.+ L_{\text{clust}} (F^{(2)}_i,y_i^{(2)}, \bm{\mu}^{(2)})\right)
\end{aligned}
\end{equation}
\begin{equation}
\begin{aligned}
    L_{\text{cross}} = \frac{1}{N} &\sum_i \left( L_{\text{clust}} (F^{(1)}_i,y_i^{(2)}, \bm{\mu}^{(2)})\right. \\
    &\left.+ L_{\text{clust}} (F^{(2)}_i,y_i^{(1)}, \bm{\mu}^{(1)})\right)
\end{aligned}
\end{equation}
\begin{equation}
    L_{\text{clust}} (F_i,y_i,\mathbf{\mu}) = -\log (\frac{\exp(-d(F_i,\mu_{y_i}))}{\sum_k \exp(-d(F_i,\mu_{k}))})
\end{equation}

$L_{\text{within}}$ represents the pixel feature vectors in one view should be closer to its cluster centers. $L_{\text{cross}}$ represents the pixel feature vectors in one view should be closer to the assigned cluster centers in the other view as well.

Note here, the K-means procedure is done after one epoch training on the whole dataset, and then its cluster center and label assignments are fixed during the training of the next epoch.  This offline and disjoint design is crucial to prevent trivial solutions.

\section{Approach}

\begin{figure*}
  \centering
  \includegraphics[width=\linewidth]{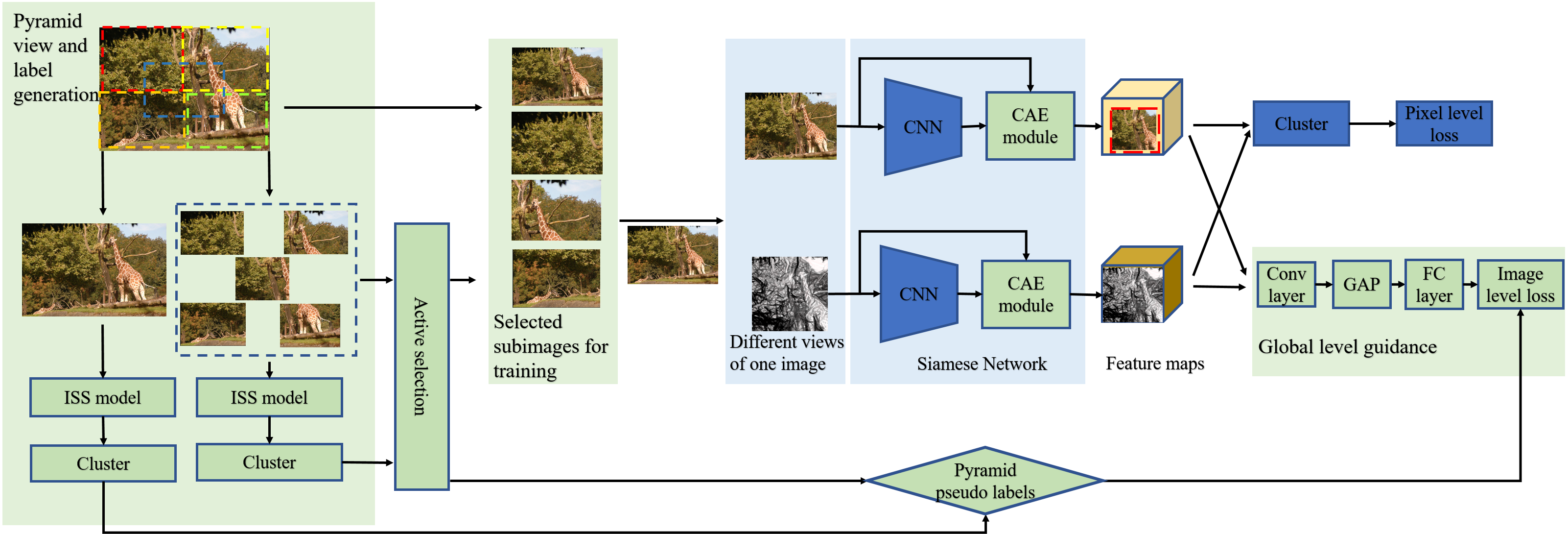}
  \caption{The overview of our framework. The \textit{green parts} are novel modules proposed in this work. The whole framework is guided by our pyramid global guidance(PGG) strategy, which includes the pyramid view and label generation, active selection, and the global-level guidance in the clustering process. In particular, the approach consists of two phases, which are our pyramid view and label generation with active selection, and the pixel clustering process. We generate a pyramid of images and assign each subimage in the pyramid of an image a pseudo label, and the subimages selected by active selection are treated as independent images for training. During the clustering process, the feature learning and clustering processes are guided by pseudo labels. Specifically, we feed each training image to two branches shown above. In the top branch, we apply geometric transformations on the feature map, while we apply the same geometric transformation on the image directly in the bottom branch. The two branches process different views but share the CNN network. Features of each view are extracted and collected for both pixel clustering and image classification. Note that the cross in the figure means each view needs to go through both the clustering and classification branches.
  }
  \label{fig:short}
\end{figure*}

\subsection{Overview}
Our fully unsupervised semantic segmentation scheme is shown in \cref{fig:short}. As a fully unsupervised scheme, the model is first trained by an instance-level self-supervised(ISS) method on a large-scale dataset. To guide the training procedure by high-level semantics, we generate pyramid pseudo labels for each image in the training set based on the ISS features using our Pyaramid-global-guided(PGG) strategy.
The pyramid pseudo labels keep unchanged and they are used as one of the supervised signals during training. We assume the global semantic acquired by the ISS model is stable and consistent among the dataset. Note that pseudo labels are generated on image pyramids to process images with multiple objects.

During training, we supervised the training based on both the generated pyramid pseudo labels and its pixel-cluster consistency\cite{picie}. In the training step, following the baseline method in section 3, we first need to generate pixel-level cluster centers and assignment labels to the dataset. We apply different photometric transformations to one image to get two views of the image. Each view is fed a Siamese network but one view is applied geometric transformations firstly. Different from the baseline\cite{picie}, to help the pixel features catch the information of the whole image, the output feature map of the Siamese network is refined by our novel CAE module, which leverages the neighborhood relationship based on both features and image raw information. Acquired the refined pixel features, we conduct pixel clustering and dataset training alternatively in a disjoint manner.

To train the network, we supervise the model with both pixel level and our image label loss. In the pixel level, we use the invariance and equivariance loss $L_p$ in Eq.(\ref{Lp-loss})

In the image level, the pyramid pseudo labels provide the loss $L_w$ of the PGG strategy in a weakly-supervision manner, where $L_w$ is a cross-entropy classification loss. This loss encourages the dominant representation to match its image pseudo label. The overall loss is formulated as follows:
\begin{equation}
    L = L_w + L_p.
\end{equation}

In the test stage, the cluster centers generated from the training data are used to label pixels in the test data. The test images are fed to the Siamese network to get feature vectors for pixels, then distances to each cluster center are computed. A post-process CRF model\cite{crf} is applied to the class scores to each cluster center to refine the segmentation.

\subsection{Pyramid-global-guided strategy}
\subsubsection{Pseudo label generation}

\paragraph{Labeling method.} We generate pseudo labels by clustering method based on an existent ISS model, such as \cite{mocov2, swav}. Specifically, we first extract image features, which are acquired by global average pooling on the feature map from the last layer of an ISS pretrained neural network. We then perform K-means on the dataset where the ISS model is trained to get cluster centers. The cluster center is then used as a classifier to assign other images. Here we normalize all the features and cluster centers to unit norm. Here note that, for some ISS methods, such as SwAV and DeepCluster, the cluster centers, or prototypes in another name, is available directly. Nevertheless, we still need to group these prototypes into smaller cluster numbers for space-saving and dataset adaptation. Given the cluster centers, an image is labeled by its nearest cluster center based on the cosine distance of image global average features. In our experiment, to compare with PiCIE\cite{picie}, we also adopt the supervised pretrained models as the initialization. In such a situation, we directly use the fully connected layer of the model to generate pseudo labels.

\paragraph{Pyramid views.} In our strategy, we apply the labeling to the pyramid of image views as shown in Fig~\ref{fig:short}, where each subimage, \ie a cropped view, in the pyramid is treated as an independent image for labeling and training afterwards.
We introduce the pyramid labeling based on the observation that a scene image could contain multiple objects and each subimage has a dominant one. One label for one image may not cover all the information in that image. This inspires us to divide the images into several subimages and assign each subimage a pseudo label. In our setting, we divide the image into 5 small crops of four on corners and one on the center of the image, then we have 6 views in all for one image, considering the original one. With the help of the ISS model, we obtain a set of pyramid labels for each image. 

\paragraph{Active selection. } To generate stable pseudo labels, we also proposed an active selection procedure to pruning the views of ambiguous semantics. Since the small subimages of images are only part of the image and its object may be incomplete. This scenario would lead to ambiguous semantic, and a lower score to its closest cluster center. In the end, we design an active selection mechanism to decide whether a label in the pyramid is used for training. We calculate the probabilities of the labels in pyramid views. In details,we assume that the distance between the global feature and the cluster center reflects the probability that the view contains distinguishable contents. Based on this assumption, we rank the view images according to their feature distance with their cluster centers, and then we only select the top 40\% images for training. Here the parameter 40\% is chosen based on experimental performance. 

Once the image pseudo labels are generated, we fix them during the whole training procedure. Global information and local pixel mining are both important for the segmentation task and they are complementary to each other. We keep the image clustering and pixel mining disjoint to prevent the stable global information from the ISS model distracted by pixel local features.

\subsubsection{Global guidance}
Given the image pseudo-labels, during training, we add a classification module after the feature map to learn the image-level information from the pseudo labels. 

Similar to class activation map (CAM)\cite{cam}, we apply a softmax function to the output feature map of the backbone to calculate the probabilities of different classes. The cross-entropy loss is computed between the pseudo labels and the probabilities. The loss function is formulated as follows:
\begin{equation}
    L_w  = - \log (\frac{\exp( q_l)}{\sum_m \exp (q_m)}), 
    \label{l1}
\end{equation}
where $q$ is output of the classification module and $q_m$ is the score on the $m$-th cluster center or class. In Eq.(\ref{l1}), $l$ is the pseudo label. In this loss, we encourage the dominant semantic pooled on the image to match a pre-defined image pseudo-label which is explicitly learned on instances.

The classical CAM~\cite{cam} compute $q$ by $q=\textbf{GAP}(g(F))$, where $g$ is a $1 \times 1$ convolution layer for classification  and $F \in \mathbb{R}^{C\times H\times W} $ is the feature map after the CAE module. GAP indicates global average pooling here. In our work, we modify this module to $q=g_1(\textbf{ReLU}(\textbf{GAP}(g_2(F))))$, where $g_1$ and $g_2$ are linear layers and $1\times1$ convolution layer respectively. The $g_1$ is for classification. We add a $\textbf{ReLU}$ layer and a linear layer after the original output. This modification makes the classification module can consider all the pixels by a non-linear function before finally deciding the category of the image. We experimentally find that the non-linear operation improves the ability of classification and summarizing. Here we use $g$ or $g_1$ to classify the image or one of the pyramid views to one of global semantics generated by PGG. Note that, the PGG and pixel-level clustering have separate cluster pools.

\subsection{Context-aware embedding module}
\begin{figure}
  \centering
  \includegraphics[width=\linewidth]{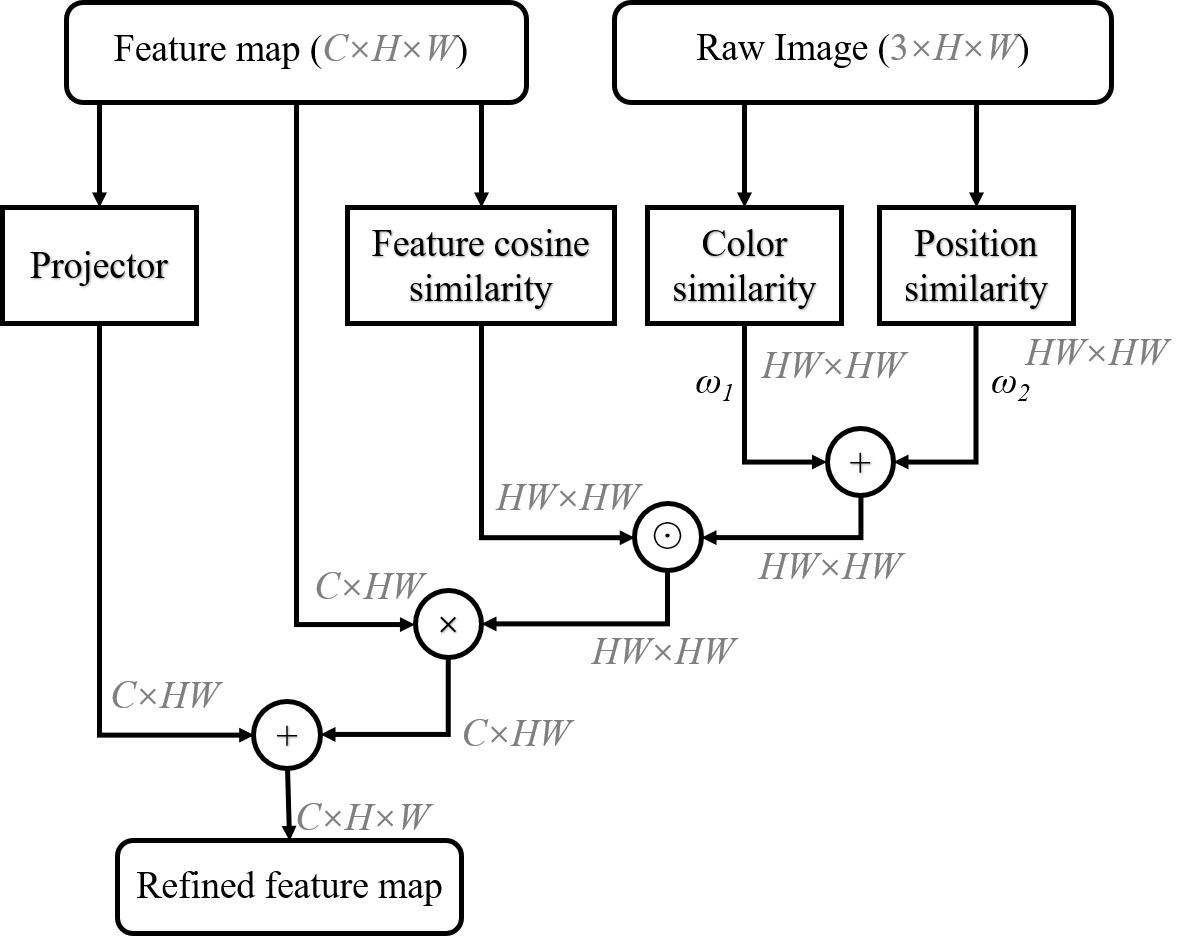}
  \caption{The structure of CAE module. The $\odot$ in the figure represents Hadamard product.}
  \label{fig:attention}
\end{figure}

To help the embedding vector of each pixel learn the knowledge of the whole image, we proposed a context-aware embedding (CAE) module which uses a self-attention mechanism. The CAE module is shown in Fig.~\ref{fig:attention}. We build the communication among pixels based on their similarities. The similarity among pixels is defined by three types of features: 1) the high-level embedding from CNN features; 2) the color feature on the raw image; 3) the geometrical feature, i.e, the spatial coordinates in the image. We calculate the cosine similarity on CNN features, that is $s_{ij} =d(f_i,f_j) = \frac{f_i \cdot f_j}{\|f_i\|\|f_j\|}$, where $f_i$ is the feature vector for pixel $i$ in feature map $f$. 

Inspired by \cite{crf}, we formulate the similarity on color and position features via a Gaussian kernel as follows.
\begin{equation}
    k_{ij} = \omega_1 k_{1ij} + \omega_2 k_{2ij},
\end{equation}
\begin{equation}
    k_{1ij} = \exp (-\frac{|p_i-p_j|^2}{2\theta_1^2}-\frac{|I_i-I_j|^2}{2\theta_2}),
\end{equation}
\begin{equation}
    k_{2ij} = \exp (-\frac{|p_i-p_j|^2}{2\theta_3^2}),
\end{equation}
where $p_i$ denotes the position of pixel $i$, $I_i$ is the color vector of pixel $i$, $\omega_1$ and $\omega_2$ are two positive linear combination weights. The $k_1$ shows that the closer distance from one pixel to the other in color space, the more attention should be paid while considering the position. The $k_2$ indicates one pixel should pay more attention to the nearer pixels. The non-local factor for pixel $j$ to pixel $i$ is then calculated as $P_{ij} = s_{ij}k_{ij}$. This procedure is also similar to a graph model\cite{crf}, in which the parameters are manually defined. In our situation, however, it is hard to pick suitable values. We set these parameters as differentiable variables and use backpropagation to find the proper values. Finally, the feature map is refined by $F_i = h(f_i) + \sum _j P_{ij}f_i  $, where $F_i$ is the new feature vector for pixel $i$ and $h$ is a projector which is a $1 \times 1$ convolution layer. In our work, we surprisedly find that encoding raw information on the graph edge is better than using CNN features only. It could be because in an unsupervised scenario CNN feature can collapse locally that the features learn to one point even though they belong to different objects and their appearances are of great difference. The raw information on relations plays a complementary role here. 

\subsection{Post process}

After the clustering procedure, we refine the segmentation to follow object boundaries leveraging a CRF model\cite{crf}. We calculate the probabilities of each pixel to the cluster centers as the unary of the graphical model. We model the edge on the graph by the color and positional information. The CRF model actually has a similar function with the proposed CAE module, but we found that an additional CRF model can still make the segmentation rough 0.5 mIoU better.

\section{Experiments}
In our method, we perform the fully self-supervised training for segmentation without any kinds of annotations. In experimental evaluation, we show comparison with previous methods, ablation study and extension evaluations on fully-supervised initialization and the semi-supervised setting. 

\subsection{Implementation details}
\paragraph{Dataset.} Here we evaluate our methods on two large-scale datasets, \ie COCO-stuff dataset\cite{cityscapes} and Cityscapes \cite{cityscapes}. The COCO-stuff dataset is a dataset containing 80 things categories and 91 stuff categories. In \cite{picie}, its classes are merged into 27 categories (15 stuff and 12 things), with 49629 for training and 2175 for testing. We follow \cite{picie} for training and testing. In the evaluation, we evaluate our method both on stuff (background contents) and things (foreground objects). Cityscapes \cite{cityscapes} is a dataset with street scenes collected from fifty cities. We process the dataset exactly the same way as \cite{picie} for fair comparison. Follow\cite{picie}, We use the same 7 out of the 8 groups for training and testing. The ground truth testing cluster number is set as 27. Its $\tt{train}$, $\tt{train\_extra} $
and $\tt{test}$ 
subsets are used for training and test on the $\tt{val}$ 
subset. To prove that our method has ability to leverage additional data, we train our model with and without additional data of a similar domain, such as the ImageNet dataset\cite{imagenet}, individually. Note that, even though we use additional data, we not use their annotations and our model is still fully self-supervised.

\paragraph{Model architectures.} We mainly adopt the ResNet-50\cite{resnet} with FPN\cite{FPN} as our backbone unless otherwise mentioned. The FPN decoder, which is consists of a convolutional layer and an upsampling module, uses the output feature map from the residual layer-1 to layer-4. the FPN output dimension is set as 256. For the CAE module, we downsample the raw image and feature map to 1/16 of the original resolution to calculate the relation map. After the relationship modeling, the refined feature map is upsampled to 1/8 of the image resolution to match the projector output which is a linear process on the FPN feature. Here we by default use our full model except that in table.~\ref{tab:iic-stuff} we use the classical CAM structure.

\paragraph{Training details.} In our work, we first pretrain a model using instance-level unsupervised methods, \ie SwAV\cite{swav} and MoCoV2\cite{mocov2}, for initialization. The model is then used for both the subsequent training and global image-level pseudo label generation by our PGG. When we train our fully self-supervised model with additional data, \ie ImageNet dataset, we just use their official released models\cite{swav,mocov2} and then perform the above-mentioned process on the target dataset. Here we use the same hyperparameters such as learning rate, pixel-level cluster number, batch size with \cite{picie} except that we set FPN dimension as 256. In our work, we perform clustering on two levels, \ie local and global, and they have separate cluster pools. For the local level, the group number is decided by their ground truth semantic class number, which is 27 on both COCO-stuff and Cityscape datasets. For the global level, we group 50 clusters by default.

\begin{figure*}
  \centering
  \includegraphics[width=0.85\linewidth]{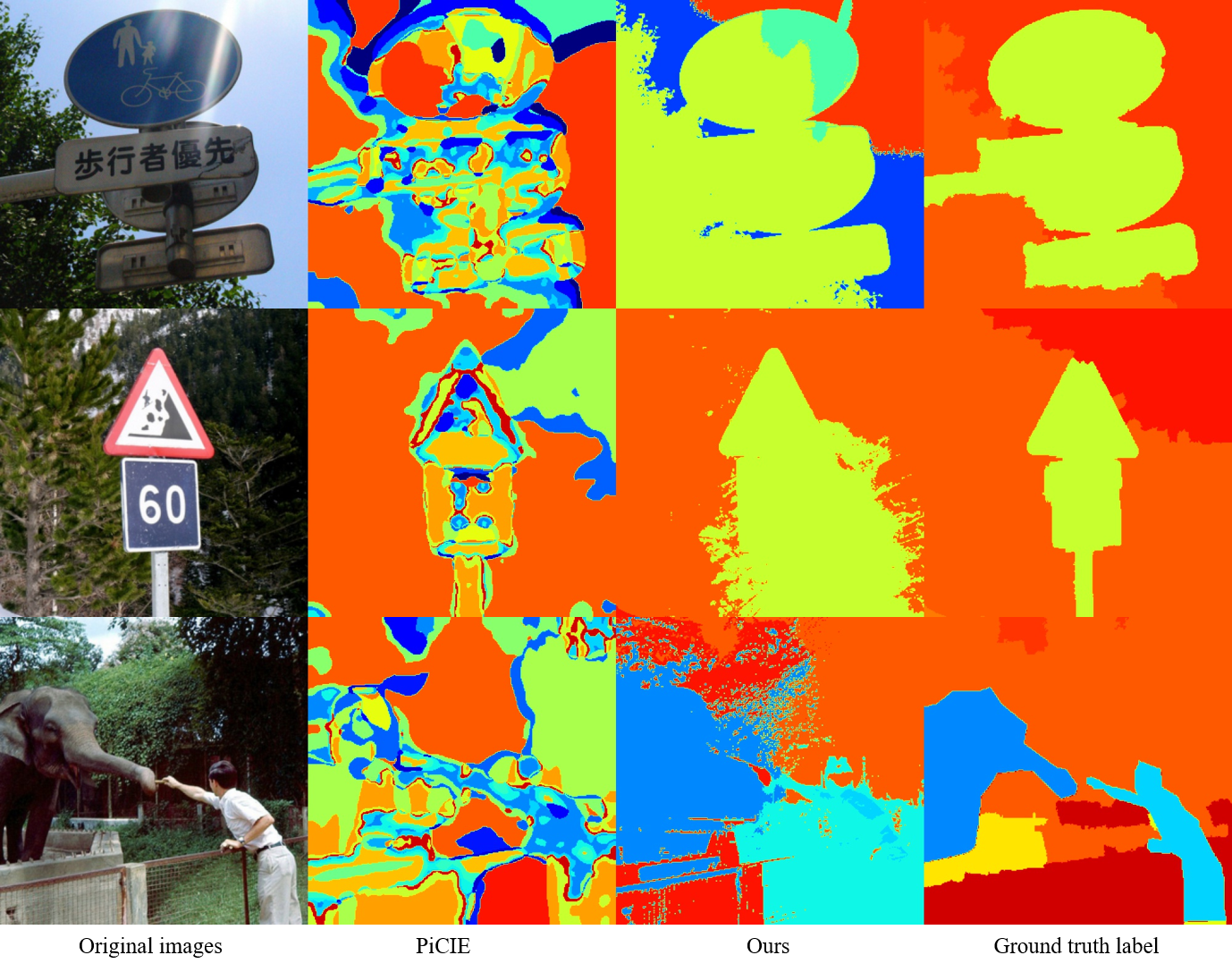}
  \caption{The results of semantic segmentation.  }
  \label{fig:vis}
\end{figure*}

\paragraph{Evaluation details.} After training, we run the trained model on the training images again and cluster the pixel-level features using the K-mean algorithm. The clustering centers are then used to segment the testing images. According to the distance between the pixel features and the cluster center, each pixel is assigned a label. Depending on the predicted segmentation, we then use Hungarian-matching\cite{assign} to match the cluster centers with the ground truth labels. Note that, for a fair comparison, all our statistical results in experiments are achieved on models without the post process. We only visualize the post process in Fig.~\ref{fig:introfig} and Fig.~\ref{fig:vis}. 

\subsection{Comparison with SOTA on $FS^4$ task}
\paragraph{Comparison on COCO-stuff.}
For the fully self-supervised semantic segmentation, there are only a few works explored on the task. Here we trained our model depending on only the COCO-stuff dataset without additional data and fully-supervised pretrained model. In particular, following our pipeline we first train an initial model using ISS methods\cite{mocov2, swav}, and then apply our PGG strategy and pixel-clustering training. We compare our methods with previous methods trained on the same setting in Table.~\ref{tab:iic-stuff}, where results on previous works are adopted from \cite{picie}. For a fair comparison, we also re-implement PiCIE by altering their initial model with a better-pretrained model, which is trained by MoCoV2 on the COCO-stuff dataset. With an ISS method pretraining, the PiCIE performs better. From the results, we can see that our method performs much better than previous works.  We outperform the better version of PiCIE by 15.5\% accuracy which attests to the superiority of our method. Here we compare accuracy results due to that most of the previous methods only report the accuracy results.

\begin{table}
  \centering
  \begin{tabular}{lc}
    \toprule
    Method &Accuracy \\
    \midrule
    Random CNN &19.4 \\
    K-means\cite{scikit-kmean} & 14.1\\
    SIFT\cite{sift}  & 20.2 \\
    Doersch 2015 \cite{Doersch}& 23.1 \\
    Isola 2016 \cite{isola}&24.3\\
    Deep cluster\cite{deepcluster} & 19.9\\
    {IIC} \cite{iic} & 27.7 \\
    AC\cite{ac}  &30.8 \\
    PiCIE\cite{picie} &31.48\\
    \midrule
    PiCIE*&33.0\\
    Ours&\textbf{48.5}\\
    \bottomrule
  \end{tabular}
  \caption{\textbf{Comparison on COCO-stuff dataset}. We compare our fully-supervised semantic segmentation model with previous works of the same setting. All the models are trained on the COCO-stuff dataset without a fully-supervised pretrained model and additional data. Here baseline results are copied from the PICIE paper. For a fair comparison, we also re-implement PiCIE by feeding it a better initial model which is trained by MoCoV2 first on the COCO-stuff dataset. This version is denoted as PiCIE* in the table. Our model performs best in this setting. }
  \label{tab:iic-stuff}
\end{table}

\paragraph{Comparison on models trained with additional data.}
To demonstrate the stability of our method and our ability of leveraging additional data which has close domains with the target scene, we also train our model based on additional data, such as the ImageNet dataset, for the COCO-stuff and Cityscapes target scenes. Here we grabbed the officially released pretrained model trained by \cite{mocov2} on the ImageNet dataset for initialization. We assume the pretrained model has contained the knowledge of new data and we can leverage the additional data through the pretrained model. We then feed the pretrained model to the PiCIE and our pipeline individually for segmentation. The results are shown in Table~\ref{tab:sota} and Table~\ref{tab:Cityscapes}.  Note that we re-implement the PiCIE with its official code. From the results, we can see that our method stably outperforms PiCIE on both category splits on different initializations. We nearly double the mIOU on \textit{All} categories and reach the upper bound shown in Table.~\cite{picie}, which demonstrates the effectiveness of our method again. For futher study, we trained our model on SwAV ImageNet pretrained model as well, and we get 13.56 mIoU while PiCIE get 6.78 mIoU. This result evidences that our work performs stable and robust on different self-supervised initialization methods.

We also show the visualization comparison with the baseline method\cite{picie}. From visualized results, we can see that our segmentation can focus on higher-level semantics, i.e. objects, and prevent segmentation of too many small regions. 

\begin{table}
  \centering
  \begin{tabular}{cccc}
    \toprule
      Method & Partition &  Acc. & mIoU \\
     \midrule
    
    PiCIE\cite{picie}   & \multirow{2}*{Things} &\textbf{47.99} &{13.89} \\
    Ours  & & {45.87} & \textbf{22.13} \\
    \midrule
    PiCIE\cite{picie}  & \multirow{2}*{Stuff} & 45.66&12.82 \\
    Ours & & \textbf{68.23} & \textbf{28.30} \\
    \midrule
    PiCIE\cite{picie}   & \multirow{2}*{All }&28.98 &7.57 \\
    Ours  && \textbf{40.38} &  \textbf{14.76} \\
    \bottomrule
  \end{tabular}
  \caption{\textbf{COCO-Stuff dataset - Comparison on models trained with additional data.} To prove our ability of leveraging additional data of similar domains, we trained the models with initialization which is trained on the ImageNet dataset with an ISS method. Note here, we still keep the model fully self-supervised since we do not use any annotations. 
  }
  \label{tab:sota}
\end{table}

\begin{table}[h]
  \centering
  \begin{tabular}{cccc}
    \toprule
    Method&\makecell[c]{ISS Pretrained\\ method}&Acc.   & mIoU  \\
    \midrule

    PiCIE\cite{picie}  &SwAV&  39.90 & 9.81\\
    Ours&SwAV&  \textbf{41.66} & \textbf{11.68}\\
    \midrule
    PiCIE\cite{picie} &MoCoV2&  \textbf{42.09} & 9.07\\
    Ours&MoCoV2&  37.66 & \textbf{10.14}\\
    \bottomrule
  \end{tabular}
  \caption{\textbf{Cityscapes dataset - Comparison on models trained with additional data.}We trained the models with initialization which is trained on the ImageNet dataset with an ISS method. }
  \label{tab:Cityscapes}
\end{table}

\subsection{Ablation study}

\paragraph{Effectiveness of different components.} We perform the ablation study on the COCO-stuff dataset by incrementally adding our modules or training operations to our model, which includes 1) additional finetune on COCO-stuff dataset with ISS methods, denoted as \textbf{COCO-stuff finetune}; 2) the \textbf{CAE module} with classical CAM structure; 3) the proposed pseudo-labeling and global guidance strategy with one label per image, denoted as \textbf{GG.}; 4) its improved version with pyramid labels for each image, i.e. our \textbf{PGG strategy}; 5) and our modified CAM structure denoted as \textbf{Mod.}.  We evaluate our method with the ResNet-50 backbone that is trained by MoCoV2 on additional data of the ImageNet dataset. The results are shown in Table~\ref{tab:moco1}. The results indicate that each of our proposed modules contributes to the final improvements, among them, our GG. module gains the most dramatic improvement. 

In addition, for the CAE module, we attempt to model the relationship with only the low-level raw information and with only the high-level CNN feature individually. On these two settings, our models achieve the mIoU of 12.01 and 9.83 respectively. It indicates that with our non-trivial design modeling raw relationships is helpful.

\begin{table}
  \centering
  \begin{tabular}{cccccc}
    \toprule
     \makecell[c]{COCO-stuff\\ finetune}& \makecell[c]{GG.}  &      \makecell[c]{CAE\\ module}& \makecell[c]{PGG\\ strategy}& Mod.  & mIoU \\
    \midrule
        
      & & & &  &7.57 \\
      \checkmark & & & &   & 7.96\\
     \checkmark &\checkmark & & &  &9.76 \\
     \checkmark &\checkmark & \checkmark&  & &12.61 \\
     \checkmark &\checkmark & \checkmark&\checkmark & & 13.43 \\
     \checkmark &\checkmark & \checkmark&\checkmark &\checkmark & 14.76 \\
    \bottomrule
  \end{tabular}
  \caption{\textbf{Ablation study - Effectiveness of different modules.} We show the results when using the MoCoV2 ISS pretrain method. COCO-stuff finetune means we use the MoCoV2 algorithm to train the model further on the COCO-stuff dataset. GG. means global guidance that we generate one pseudo label for each image for image-level supervision and do not use the pyramid views for training. }
  \label{tab:moco1}
\end{table}

\paragraph{Study on global cluster number.} In Tab.~\ref{tab:moco2}, we show how the class numbers in our GG strategy influence the results. The results present that our method is insensitive to the cluster/class number. The performances are stable on a set of cluster numbers before 100. It may indicate that a cluster number close to the true class number of the dataset, that is 27 for the COCO-stuff dataset, can give better results. 

\begin{table}
    \centering
    \begin{tabular}{ccccc}
        \toprule
        \#label & 25 &50&100&200 \\
        \midrule
        mIoU & 13.11 &12.61&12.40&10.55\\
        \bottomrule
      \end{tabular}
      \caption{\textbf{Ablation study on global cluster number.} For image-level supervision when using MoCoV2 initialization on COCO-stuff dataset. }
      \label{tab:moco2}
\end{table}

\subsection{Comparison on fully-supervised initialization.} To compare with PiCIE, we also follow its setting using ResNet-18 as the backbone and initializing the training from the fully-supervised pretrained model on ImageNet\cite{imagenet}. Results are shown in Table~\ref{picie}. Although our method is designed for a totally unsupervised manner, we can still outperform the prior art, which indicates the global guidance is still helpful when the learning starts from a good state. We produce the pseudo labels for training images from the fully connected classification layer of the supervised models. The results here serve as an upper bound for our fully unsupervised setting. We can see that our method bridges the gap raised by suboptimal initialization and its result nearly reaches the upper bound. 

\begin{table}
  \centering
  \begin{tabular}{lc@{}}
    \toprule
     Method & mIoU \\
    \midrule
     PiCIE+H.& 14.36  \\
     Ours+H. & \textbf{15.69}\\
    \bottomrule
  \end{tabular}
  \caption{\textbf{Compassion on fully-supervised initialization.}
  To compare with the PiCIE, we add the same over-clustering loss as described in \cite{picie}, which is denoted as +H.  
  }
  \label{picie}
\end{table}

\section{Conclusion}
In this paper, we proposed a novel method that is able to train a segmentation model in a fully unsupervised manner. The proposed pyramid global guidance(PGG) strategy and context-aware embedding(CAE) module encourage pixel features to pay attention to high-level image semantics while learning their own concept. We show that our method can effectively and automatically discover high-level semantics without any human labels. The method we propose is also robust to the different training starting points.
Nevertheless, unsupervised semantic segmentation, due to lacking strong and stable supervision signal, can easily run into local collapsing, we will keep making effort to solve this problem in the future.

{

}
\clearpage
\begin{appendix}
    \section{Experiment details}
    The base training settings closely follow PiCIE\cite{picie}. We describe these parameters as follows. 
    \subsection{Data processing}
    During training, we pre-process the images by resizing and center cropping to $320\times  320$. We apply random photometric transformations including color jitter, gray scale, and Gaussian blur with the probability 0.8, 0.2, 0.5. The color jitter contains jittering brightness, contrast, saturation and hue, whose control factors are 0.3, 0.3, 0.3, 0.1.
    The control factor of Gaussian blur is randomly chosen from [0.1, 2.0]. For geometric transformation, we use random crop and random horizontal flip. The crop scale is randomly chosen from [0.5, 1.0]. The probability for the horizontal flip is 0.5.
    
    When assigning the pseudo labels, we follow \cite{swav,mocov2} to resize images to $256\times 256$ and center crop them to $224\times 224$ for inference. We then use the features from projectors for clustering. For pyramid pseudo labels, we crop five views for each image from its four corners and the center. These views are resized to $640\times 640$ and then processed following the same procedure for image-level pseudo label generation.
    
    \subsection{Training parameters}
    We train the network for 10 epochs using ADAM optimizer with the learning rate 1e-3 and no weight decay. The batch size is 128.
    
    In our experiment, we find out that unbalanced loss leads to a better result for PiCIE\cite{picie}. The result of unbalanced loss is 6.78 mIoU, while the result of balanced loss is 6.23, when PiCIE method uses SwAV\cite{swav} initialization. We thus report the results on unbalanced loss in the main paper.
    \section{Visualization}
    \subsection{Segmentation comparison}
\begin{figure*}
      \centering
      \includegraphics[width=0.8\linewidth]{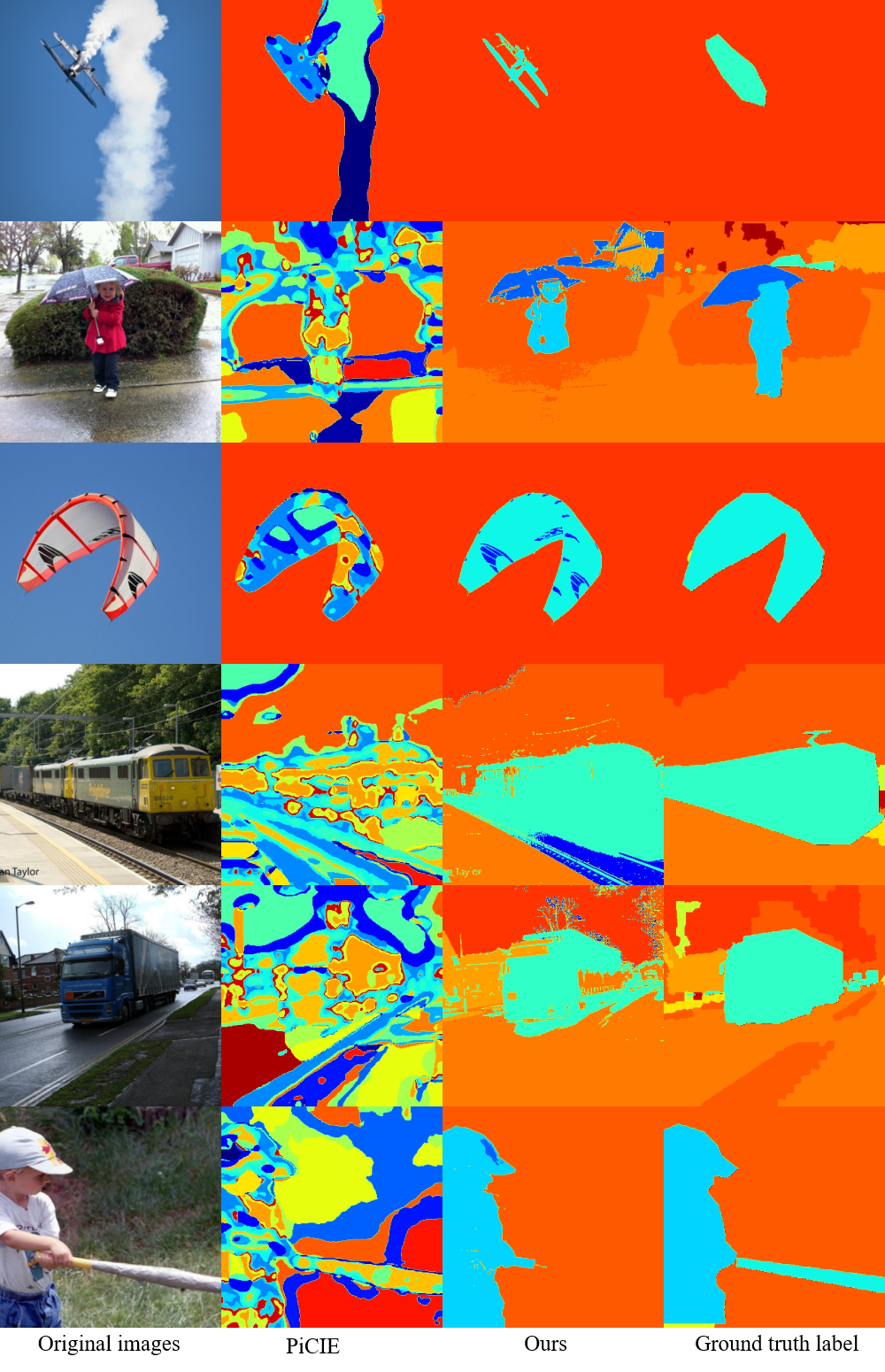}
      \caption{Segmentation results.}
      \label{fig:suppseg1}
    \end{figure*}

    Here we present more segmentation results in Fig.~\ref{fig:suppseg1}. Our results are processed with our full model with the CRF\cite{crf}. From these figures, we can see that our method can better understand the high-level semantics of images, while for the foreground objects PiCIE\cite{picie} is more concentrated on the low-level cues such as the edges and the colors.
    
    Since our method leverages the global pseudo labels, we pay much attention on dominant objects and scene stuff, and in some cases neglect small objects in the scene. This issue can be solved to some extent by our pyramid labels. We will also take it as our future work to solve this issue.
    
    \subsection{Image-level pseudo labels}
\begin{figure*}
      \centering
      \includegraphics[width=\linewidth]{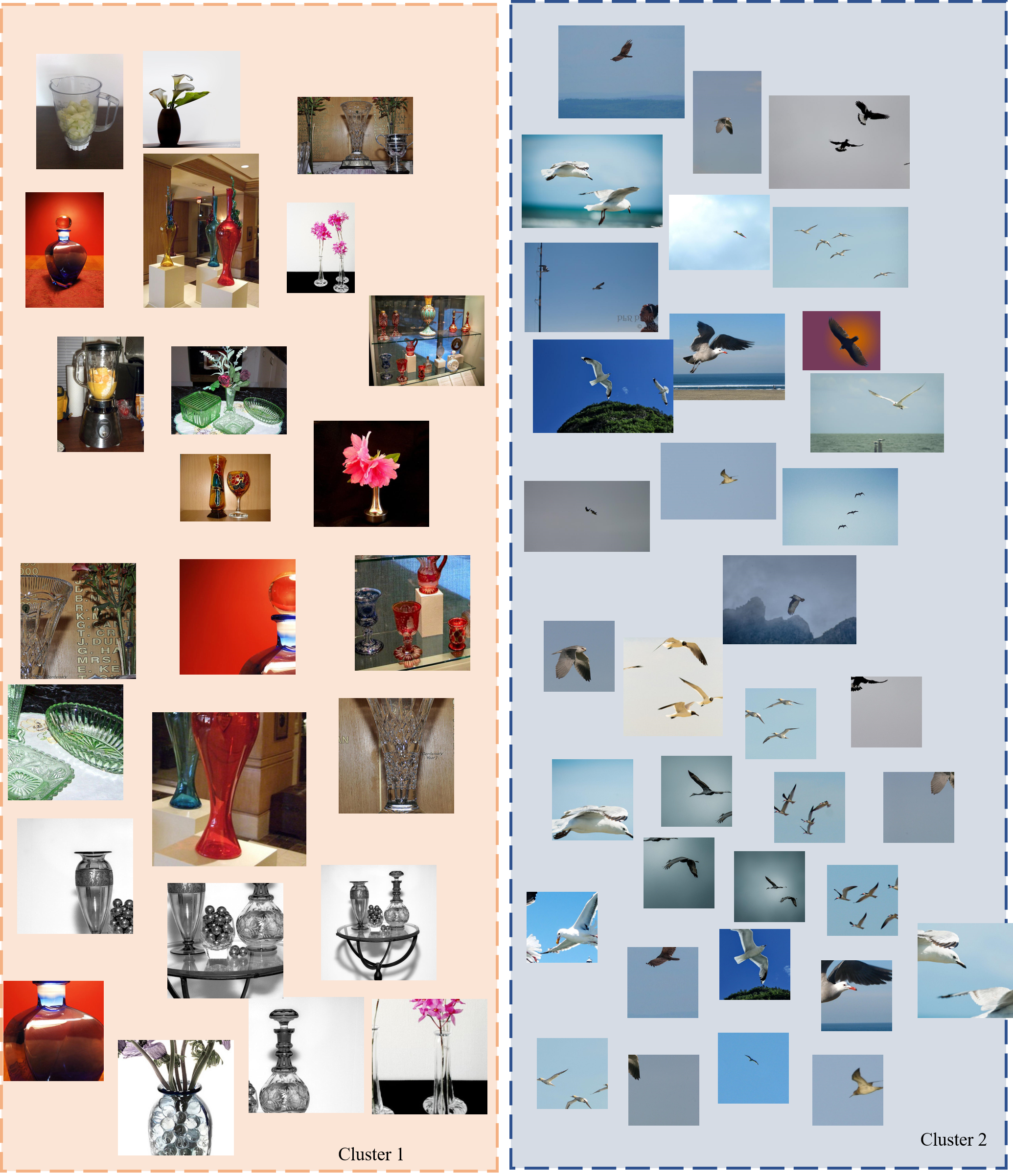}
      \caption{Image grouping results. We show the images and crop views in cluster 1 and cluster 2. Cluster 1 mainly contains images with glass. Cluster 2 mainly contains images with bird. The crop views are more concentrated on these objects. }
      \label{fig:clu1}
    \end{figure*}

    We present the images with the same pseudo labels in Fig.~\ref{fig:clu1}, where the images are from both the original views and crop views. From the visualization, we can conclude that our pseudo label generation strategy can effectively group the images with similar semantics together. 
    
\end{appendix}

\end{document}